\pdfoutput=1
\documentclass{llncs}
\usepackage{graphicx}
\usepackage{subcaption}
\usepackage{amsfonts}
\usepackage{tikz,graphicx}
\usepackage{xcolor}
\usepackage{hyperref}
\usepackage{cleveref}
\usepackage{booktabs}
\usetikzlibrary{positioning}

\hypersetup{
  pdfauthor={Yizi Chen, Edwin Carlinet, Joseph Chazalon, Clément Mallet, Bertrand Duménieu, Julien Perret},
  pdftitle={Combining Deep Learning and Mathematical Morphology for Historical Map Segmentation},
  pdfkeywords={Deep Learning, Convolutional Neural Networks, Mathematical Morphology, Historical Map Segmentation, Object Extraction}
  }

\newcommand\blfootnote[1]{%
  \begingroup
  \renewcommand\thefootnote{}\footnote{#1}%
  \addtocounter{footnote}{-1}%
  \endgroup
}

\begin{document}
\title{Combining Deep Learning and Mathematical Morphology for Historical Map Segmentation}

\author{%
Yizi Chen\inst{1,2} \and 
Edwin Carlinet\inst{1}\and
Joseph Chazalon\inst{1}\and\\
Clément Mallet\inst{2}\and
Bertrand Duménieu\inst{3}\and
Julien Perret\inst{2,3}%
}

\authorrunning{Y. Chen et al.}
\institute{%
EPITA Research and Development Lab. (LRDE), EPITA, France \and
Univ. Gustave Eiffel, IGN-ENSG, LaSTIG \and
LaDéHiS, CRH, EHESS}
\maketitle              %
\begin{abstract}
The digitization of historical maps enables the study of ancient, fragile, unique, and hardly accessible information sources. Main map features can be retrieved and tracked through the time for subsequent thematic analysis.
The goal of this work is the vectorization step, i.e., the extraction of vector shapes of the objects of interest from raster images of maps.
We are particularly interested in closed shape detection such as buildings, building blocks, gardens, rivers, etc. in order to monitor their temporal evolution.
Historical map images present significant pattern recognition challenges.
The extraction of closed shapes by using traditional Mathematical Morphology (MM) is highly challenging due to the overlapping of multiple map features and texts.
Moreover, state-of-the-art Convolutional Neural Networks (CNN) are perfectly designed for content image filtering but provide no guarantee about closed shape detection.
Also, the lack of textural and color information of historical maps makes it hard for CNN to detect shapes that are represented by only their boundaries.
Our contribution is a pipeline that combines the strengths of CNN (efficient edge detection and filtering) and MM (guaranteed extraction of closed shapes) in order to achieve such a task.
The evaluation of our approach on a public dataset shows its effectiveness for extracting the closed boundaries of objects in historical maps.

\end{abstract}

\section{Introduction}
\begin{figure}[tb]
  \begin{subfigure}[t]{.39\linewidth}
      \centering
      \includegraphics[trim=1cm 0 0 2.4cm, clip, width=1\textwidth]{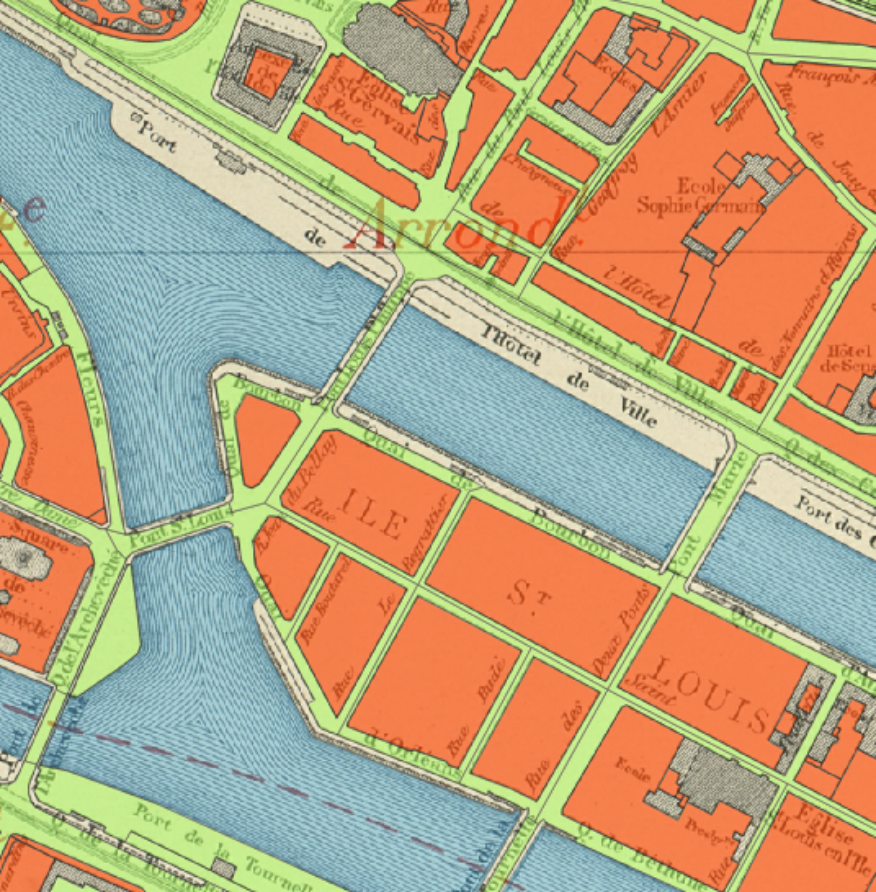}
      \caption{Some geographical entities typically depicted  in city maps: building blocks (orange), roads (green) and rivers (blue).}
      \label{fig:objects}
  \end{subfigure}
  \hfill
  \begin{subfigure}[t]{.59\linewidth}
      \centering
      \begin{tikzpicture}
      \node (img) {\includegraphics[width=\textwidth]{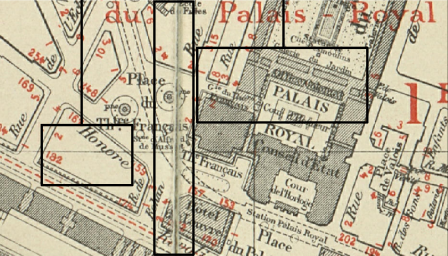}};
    \node[inner sep=0pt] at (-2.3, 2.3){(1)};
    \node[inner sep=0pt] at (1, 2.3){(2)};
    \node[inner sep=0pt] at (-0.8, 2.3){(3)};
    \end{tikzpicture}
      \caption{Challenges in historical maps: (1) planimetric overlap, (2) text overlap, (3) paper folds.}
      \label{fig:mapissues}
  \end{subfigure}
  \caption{Contents of a 1925 urban topographic map along with an overview of their challenging properties for automatic feature extraction.}%
  \label{fig:mapoverview}
  \end{figure}
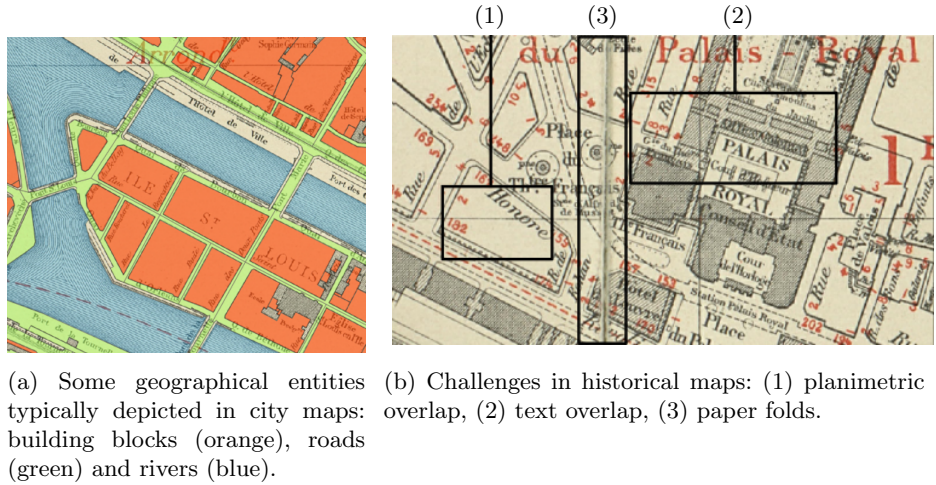
\blfootnote{%
  \scriptsize
  Extra material for this paper (full-size figures, results, code, dataset) available at:\\
  \url{https://github.com/soduco/paper-dgmm2021}
}%
The massive digitization of archival collections carried out by heritage institutions provides access to huge volumes of historical information encoded in the available documents.
Among them, maps are unfortunately still little exploited.
Yet they are a gold mine of geographic data that allows to reconstruct and analyze the morphological and social evolution of a place over time~\cite{perret2015roads,dietzel2005spatio}.
Topographic maps, in particular, engrave many geographical features: their distribution in space, their topological relationships and various information encoded by the map legend or by text labels~\cite{leyk2006saliency,chiang2011efficient}. 
Transforming such graphical representations of geographic entities into discrete geographic data (or vector data) is a crucial step for numerous spatial and spatio-temporal analysis purposes.
Such a transformation is most often manually retrieved by historians or with the help of crowdsourcing tools.
This is extremely time-consuming, non-reproducible, and leads to heterogeneous data quality.
Automating this tedious task is a key step towards building large volumes of reference geo-historical data.%

Unfortunately, historical maps exhibit characteristics that hinder standard pattern recognition approaches and make them relatively inefficient at extracting data of good quality, i.e., that do not need to be manually post-processed.
Unlike modern computer-generated maps which follow roughly the same semiotic rules, these maps vary in terms of legend, level of generalization, type of geographic features and text fonts~\cite{leyk2006saliency}.
They also usually lack texture information, which creates ambiguities in the detection of objects.  
For instance, building blocks and roads have very similar textures despite being of completely different nature (\Cref{fig:objects}).
Popular semantic~\cite{long2015fully,badrinarayanan2017segnet,ronneberger2015u} and instance~\cite{chen2019hybrid,romera2016recurrent,CVPR17} image segmentation algorithms detect objects based on textures and are prone to fail in our context.
Color is not a relevant cue either: the palette is usually highly restricted due to the technical limitations and financial constraints of their production.
Objects in maps are often overlapping, some are thus partially hidden and hardly separable.
Occlusion happens with overlayed textual and carto-geodetic information in particular (\Cref{fig:mapissues}, rectangles (1) and (2)).
Last, preservation conditions of historical maps play a role as stains, folds or holes might cause gaps in the cartographic information.
Such artifacts may lead to incorrect object detection (\Cref{fig:mapissues}, rectangle (3)).

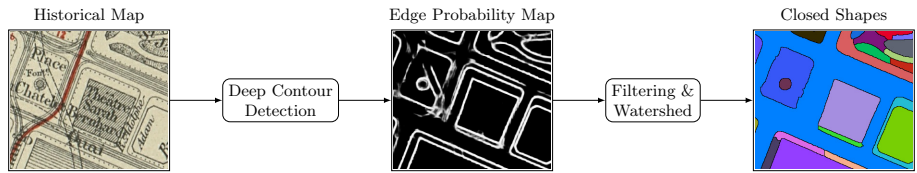
\begin{figure}[tb]
  \centering

  \resizebox {\textwidth} {!} {  
  \begin{tikzpicture}

    \tikzset{
      Image/.style={draw, inner sep=1pt},
      Process/.style={draw, rounded corners, align=center}
    }
  
    \node[Image,label={Historical Map}]                     (A1) {\includegraphics[width=3cm]{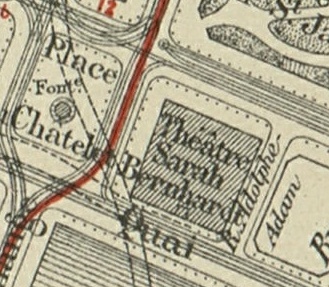}};
  
    \node[Process] [right=of A1] (A2) {Deep Contour\\ Detection};
  
    \node[Image,label={Edge Probability Map}] [right=of A2] (B1) {\includegraphics[width=3cm]{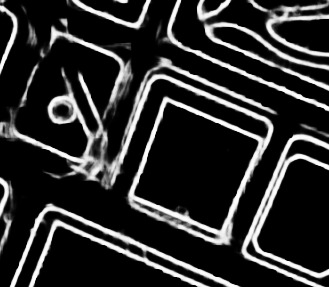}};
  
    \node[Process] [right=of B1] (B2) {Filtering \&\\ Watershed};
  
    \node[Image,label={Closed Shapes}]        [right=of B2] (C1) {\includegraphics[width=3cm]{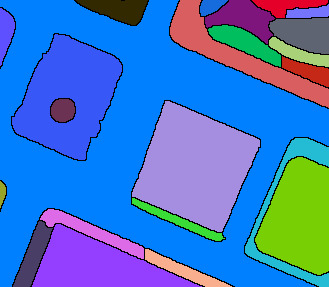}};
  
    \begin{scope}[-latex]
      \draw (A1) -- (A2);
      \draw (A2) -- (B1);
      \draw (B1) -- (B2);
      \draw (B2) -- (C1);
    \end{scope}

  \end{tikzpicture}
  }

\caption{Overview of the approach presented in the paper:
  we combine an efficient edge detection and filtering stage using a deep network
  with a fast closed shape extraction using mathematical morphology tools.}
  \label{fig.pipeline}
\end{figure}

Our contributions in this paper are as follows.
After reviewing the limitations of the current approaches for segmenting maps in \cref{sec:approach}, we
propose a simple pipeline (\cref{fig.pipeline}) that combines deep networks and mathematical morphology for object
detection in maps. It takes benefit from their complementary strengths, namely image filtering and strong guarantees
with respect to closed shapes. We derive edge probability maps using a multi-scale deep network approach
depicted in \cref{sec:deep_edge_det} and then leverage mathematical morphology tools to extract closed shapes
as explained in \cref{sec:mm_extraction}.
Eventually, in \cref{sec:eval}, the second contribution lies in a thorough evaluation of the relevance of the
mathematical morphology stage with novel visualizations and metrics to objectively assess our approach and better
identify the strengths and weaknesses of each stage and of the workflow.

\section{Approaches for map segmentation}
\label{sec:approach}

We target to recover geometric structures from scans of historical maps. As mentioned above, due to the limited texture and colour content of such data sources, standard semantic segmentation approaches of the literature would fail for most cases. Instead, we cast our problem as a vectorization challenge that can be turned into a region-based contour extraction task. Such a problem is traditionally solved through a two-step approach: the detection of edges or local primitives (lines, corners) followed by the retrieval of structures based on global constraints \cite{CVPR12}. Recent works have shown the relevance of a coupled solution \cite{PAMI_favreau}. They remain tractable and efficient only for a limited number of structures. Region-based methods (e.g., based on PDEs \cite{OBWBTS08}) may lead to oversimplified results and will not be further analysed here.\\
The main issue of two-step solutions is the edge detection step. This low-level task is achieved by measuring locally pixel gradients. Due to the amount of noise (overlapping objects, map deformation), this would result in many tiny and spurious elements that any global solution would manage connecting. Instead, we focus on boundary detection, i.e.,      
a middle-level image task that separates objects at the semantic level according to different geometric properties of images. This offers two main advantages: (i)~a limited sensitivity to noise in maps and (ii)~the provision of more salient and robust primitives for the subsequent object extraction step. We do not focus on a primitive-based approach since shapes on maps cannot be simply assumed. \\

Recently, among the vast amount of literature, convolution neural networks (CNN) have shown a high level of performance for boundary detection \cite{xie2015holistically,he2020bdcn}. However, they only provide probability edge maps. Without topological constraints, image partitioning is not ensured. Conversely, watershed segmentation techniques in mathematical morphology can directly extract closed contours. They run fast for such a generation, but may lead to many false-positive results. Indeed, using only low-level image features such as image gradients, watershed techniques may not efficiently maintain useful boundary information \cite{CVPR17}. Consequently, we propose here to merge the CNN-based and watershed image segmentation methods in order to benefit from the strengths of both strategies \cite{Neurocomputing_MMDL}. A supervised approach is conceivable since we both have access to reference vectorized maps and CNN architectures pre-trained with natural image.\\

\section{Deep Edge Detection}
\label{sec:deep_edge_det}
We detail how we selected the network architecture used to detect and filter edges,
with illustrations of the strengths of such approach, 
and describe the training procedure we followed to use the selected network (BDCN) on our dataset.

\textbf{Network architecture} %
Contour detection was first addressed with the design of handcrafted features based on brightness, color, textures~\cite{MartinFTM01}. Then, improvements lied in their efficient group  through mono- or multi-scale attributes retrieving micro-structures: textons are a salient example \cite{Zhu}. Afterwards, main methods focused on combining all available cues, such as \cite{arbelaez2010contour}. They used a global probability boundary by learning the weights of manually selected features (gradients and textons as features in several image scales) in order to detect contours and form better closed boundaries to represent the objects in images. Since CNNs have proved their relevance to extract and combine meaningful image features, a large amount of research has focused on detecting contours.
The most famous one is the so-called Holistically edge detector (HED)~\cite{xie2015holistically}, which is an end-to-end multi-scale deep learning network. 
The novelty consisted in using skip-connections to merge different levels of features and learn different losses from intermediate layers of VGG-16~\cite{simonyan2014very}. This allowed to recover multiscale representations of image features.
Eventually, He et al.~\cite{he2020bdcn} proposed a so-called Bi-Directional Cascade Network (BDCN) by designing a scale enhancement module (SEM) on top of HED to enhance multiscale spatial contexts in images resulting in a better performance than humans in the BSDS500 dataset.

\newcommand{\figwidth}{0.24\textwidth}

\begin{figure}[tb]
  \includegraphics[trim={0 6cm 0 6cm},clip,width=0.49\textwidth]{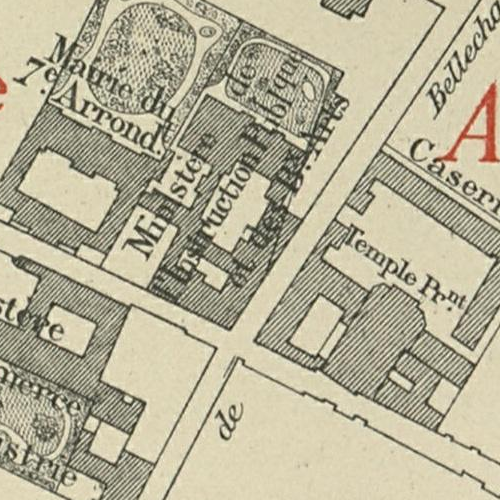}\hfill
  \hfill
  \includegraphics[trim={0 6cm 0 6cm},clip,width=0.49\textwidth]{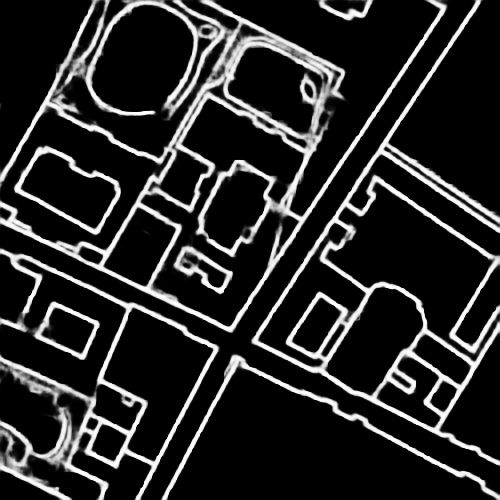}
  \caption{BDCN produces an Edge Probability Map (\emph{right}) with texts and textures removed from the input
    (\emph{left}).}%
  \label{fig.epm}
\end{figure}

One advantage of BDCN is that the multiscale representatives combine semantically meaningful features to efficiently filter out the image textures and text information while maintaining useful contours and lines in the images. 
It is particularly suited for handling noise in our maps.
Another advantage is that learnable dilated convolutions in SEM can learn fine-grained features with larger receptive fields that are beneficial when we want to separate the texts with object contours accurately.
It is because building contours have much longer
pixel continuity than text, resulting in higher activation after dilated convolution. 
After several iterations,
the probability of text pixels will vanish, leading to their removal, similarly to texture, as shown in \cref{fig.epm}. 
However, the BDCN network works only at the pixel level and cannot guarantee the required topological properties in predicted edge probability maps without additional topological constraints~\cite{clough2019explicit}, thus the current solution requires knowledge of the number of structures to be retrieved.

\textbf{Training} %
Annotated historical maps are used to train a BDCN network.
The final prediction which is a probability map where each pixel in the maps contain values in range $[0,1]$ (zero means the pixel does not belong to a contour, one that it does).
We train our network from scratch instead of using transfer learning on the edge weights learned from BSDS500 (dataset developed for image boundary detection and segmentation tasks): the features in natural images are very different from our historical map images.%
We need to filter out most of the texts in our maps, but the network trained on the BSDS dataset does not provide any useful features related to geometric filtering tasks.
In order to handle data imbalance during training, we proceed as follows.
We define our input image as $x \in$ $\mathbb{R}^{H \cdot W}$ and ground truth label $y \in$ ${0, 1}^{H \cdot W}$. 
The output of predicted image is $\hat{y} = f(x, w) \in {0, 1}^{H \cdot W}$ and every element of $\hat{y}$ is interpreted as the probability of pixel i having label 1: $\hat{y} \equiv p (Y_i = 1 | x, w)$.
Since the edge detection is a binary classification task, binary cross entropy loss is used as loss function between predictions and ground truths.
Due to highly imbalanced edge (97.5\%) and non-edge (only 2.5\%) classes, extra parameters $\alpha, \beta$ are used as weights to re-balance the binary cross entropy loss, as
$\mathcal{L}_{BCE}=-\alpha\sum_{j\in Y_{-}} log(1-\hat{y_j}) - \beta \sum_{j \in Y_{+}} log(\hat{y_j})$
where
$Y_{+}$ is the set of indices of edge pixels,
$Y_{-}$ is the set of indices of non-edge pixels,
$\alpha=\left(\lambda \cdot |Y_{-}|/(|Y_{+}| + |Y_{-}|) \right)$
is the percentage of edge pixels in each batch of historical map image 
and $\beta=\left(|Y_{+}|/(|Y_{+}| + |Y_{-}|)) \right)$
is percentage of non-edge pixels. 
An extra $\lambda=1.1$ factor is used to enhance the percentage of edge pixels in order to give extra weights for edge responses.
\\%
We build our code based on the BDCN code repository %
to train our historical map dataset from scratch with a few modifications.
We evaluate the loss for every epoch and also for choosing the best training weights. 
To make the network converging faster, we replace SGD with ADAM optimizer.
The initial learning rate is set to $5\times 10^{-5}$ with $0.9$ momentum and $0.002$ weight decay.

\section{Segmentation of the EPM}
\label{sec:mm_extraction}

From the Edge Probability Map, we then need to extract boundaries of the objects. In Mathematical Morphology, the
Watershed Transform~\cite{meyer1994topographic} is a de-facto standard approach for image segmentation. It has been
used in many applications and has been widely studied in terms topological
properties~\cite{cousty2009watershed,roerdink2000watershed}, in terms of algorithms and in terms on computation
speed~\cite{roerdink2000watershed,couprie2005quasi}.

It has two well-known issues: the over-segmentation due to the high number of minima, and the gradient leakage that
merges regions. There is a third general issue with the watershed that concerns the separation of overlapping or
touching objects, but this is not a problem in our case since the map components do not overlap.

\textbf{Solutions to the over-segmentation problem} The first problem is generally solver by filtering the minima
first. In \cite{soille2013morphological}, the \emph{h}-minima characterize the importance of each local minimum through
their \emph{dynamic}. When flooding a basin, it actually refers to the water elevation required to merge with another
basin. Attributes filter, filters by reconstruction~\cite{salembier1995flat} also allow to eliminate some minima based
on their algebraic properties: size, shape, volume\dots{} Another efficient approach consists in first ordering the
way the basins merge to create a hierarchy of partitions and then performing a cut in the hierarchy to get a
segmentation with non-meaningful basins removed~\cite{beucher1994watershed,perret2019removing,barcelos2019exploring}.

\textbf{Solutions to the early leakage problem} The second problem lies in the quality of the gradient. It has been
noted~\cite{perret2017evaluation}, that (hierarchical) watersheds have better results on non-local supervised gradient
estimators. The idea of combining the watershed with high performance contour detector dates back
to~\cite{arbelaez2010contour}.

The relevance of a simple closing by area and dynamic on the edge map produced by our deep-learning edge detector
combined with the watershed for this application lies in three points.

First, the minimum size of the components is known. Indeed, the document represents a physical size, and regions whose
area is below 100 $m^2$ are not represented in the map. Thus, we have a strong \emph{a priori} knowledge we want to inject in
the process, the minimum size of the regions (in pixels). This type of constrain is hard to infer in a deep-learning
system and we cannot have such guarantees from its output. Having hard guaranties about the shapes and their size is at
the foundation of the granulometries in Mathematical Morphology. Moreover, the connected (area) filter used for filtering
the edge image ensure that we do not distort the signal at the boundaries of the meaningful regions.

Second, the watershed segmentation method does not rely on the strength of the gradients to select the regions. Even if the
edge response is low (i.e, the gradient is weak), the watershed is able to consider this weak response and closes the
contour of the region. We do not depend on the strength of the edge response from BDCN which is difficult to calibrate
and normalize.

Last but not least, not only the watershed outputs a segmentation but some implementations also produce watershed lines
between regions. In our application, watershed lines are even more important than regions because we need to extract
polygons for each shape. Event if we could extract boundaries from regions, it avoids an extra processing step. The
watershed lines produced by the algorithm is one pixel-large and are located where the edges are the strongest, i.e.,
where the network has the strongest response on thick edges. The watershed lines form closed boundaries around regions
which is a guarantee we cannot have from the output of a network.

\begin{figure}[tb]
  \setlength{\fboxsep}{0.5pt}

  \fbox{\includegraphics[trim={0cm 2cm 0cm 0cm},clip,width=2.3cm]{./images/ws/input-crop-1.jpg}}
  \fbox{\includegraphics[trim={0cm 2cm 0cm 0cm},clip,width=2.3cm]{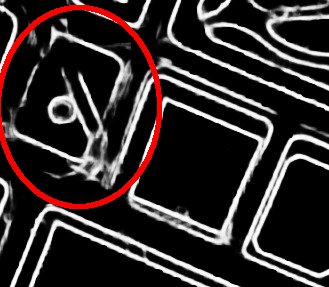}}
  \fbox{\includegraphics[trim={0cm 2cm 0cm 0cm},clip,width=2.3cm]{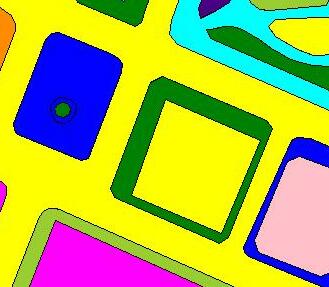}}
  \fbox{\includegraphics[trim={0cm 2cm 0cm 0cm},clip,width=2.3cm]{./images/ws/B-crop-1.jpg}}
  \fbox{\includegraphics[trim={0cm 2cm 0cm 0cm},clip,width=2.3cm]{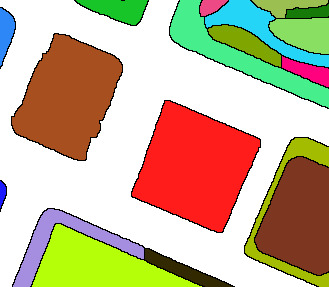}}

  \fbox{\includegraphics[trim={0cm 1cm 0cm 3cm},clip,width=2.3cm]{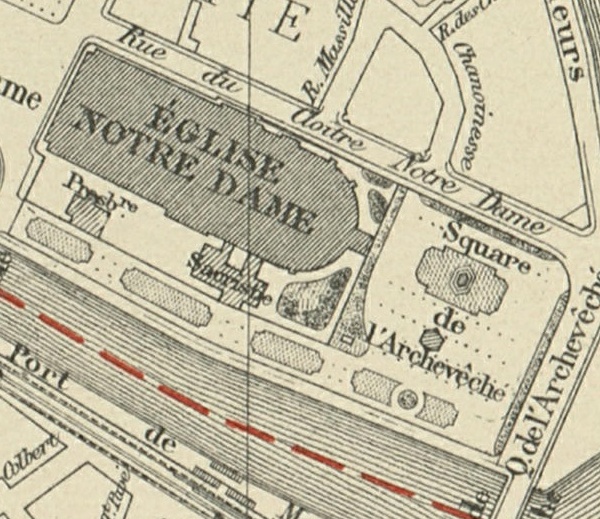}}
  \fbox{\includegraphics[trim={0cm 1cm 0cm 3cm},clip,width=2.3cm]{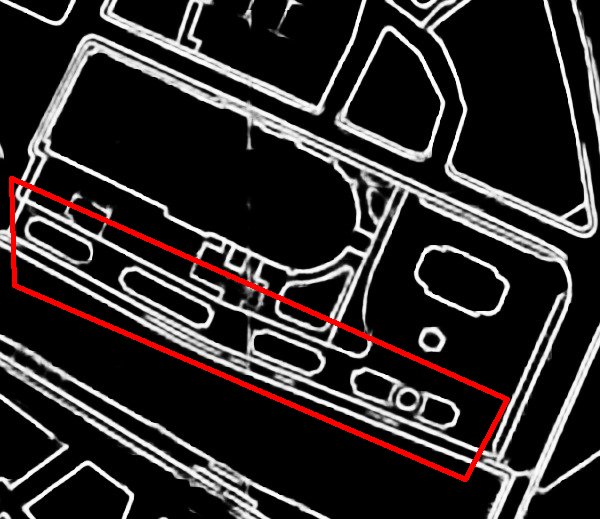}}
  \fbox{\includegraphics[trim={0cm 1cm 0cm 3cm},clip,width=2.3cm]{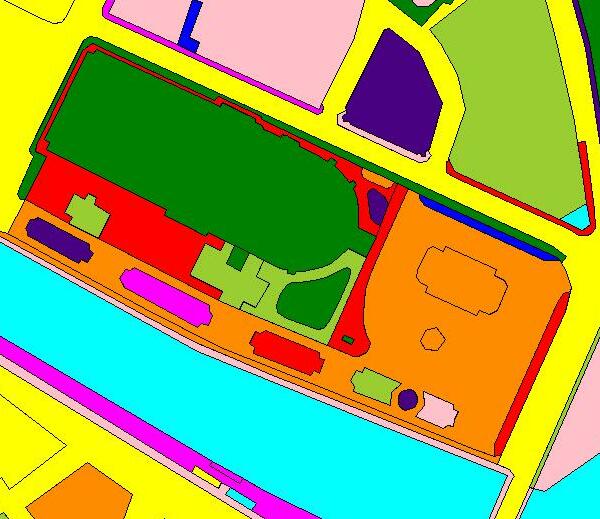}}
  \fbox{\includegraphics[trim={0cm 1cm 0cm 3cm},clip,width=2.3cm]{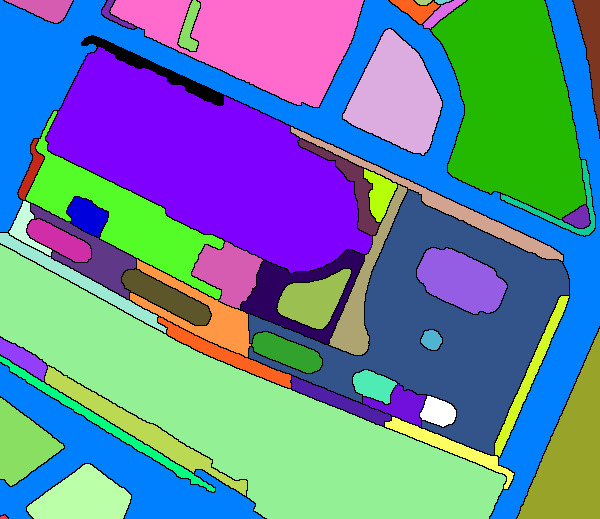}}
  \fbox{\includegraphics[trim={0cm 1cm 0cm 3cm},clip,width=2.3cm]{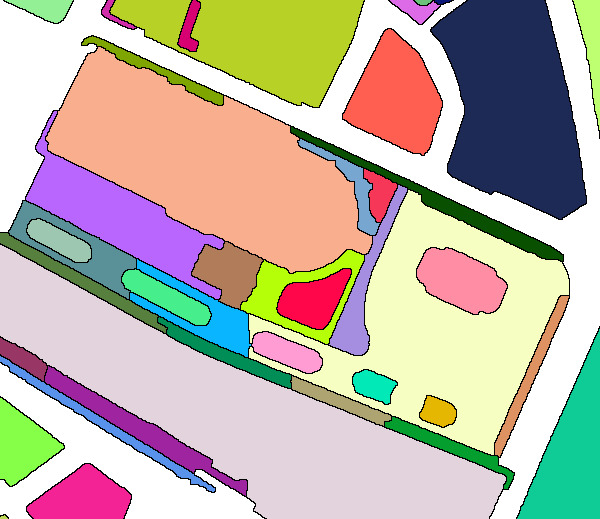}}

  \fbox{\includegraphics[width=2.3cm]{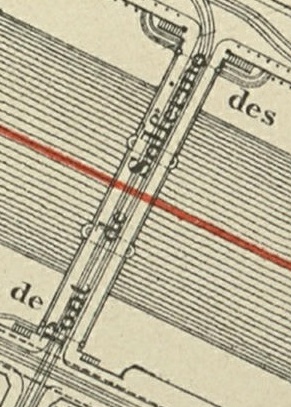}}
  \fbox{\includegraphics[width=2.3cm]{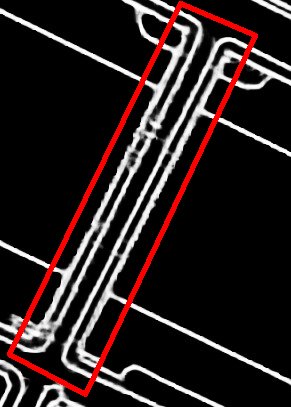}}
  \fbox{\includegraphics[width=2.3cm]{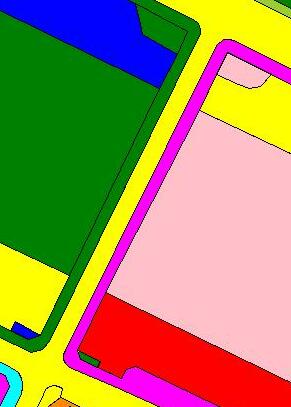}}
  \fbox{\includegraphics[width=2.3cm]{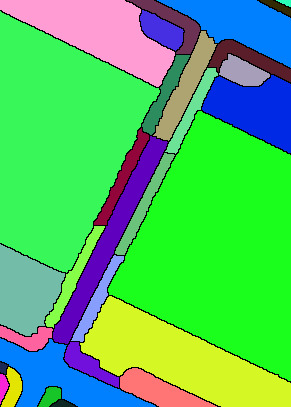}}
  \fbox{\includegraphics[width=2.3cm]{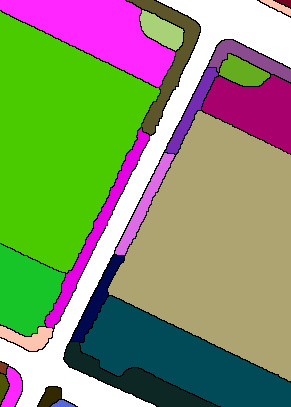}}

  \centerline{
  \makebox[2.3cm][c]{Input}
  \makebox[2.3cm][c]{EPM}
  \makebox[2.3cm][c]{Ground-truth}
  \makebox[2.3cm][c]{Param set A}
  \makebox[2.3cm][c]{Param set B}
  }

  \caption{Some \emph{failures} and some \emph{success stories} of the watershed segmentation. The parameter sets are A:
    \emph{h=3}, $\lambda=250$, and B: \emph{h=7}, $\lambda=400$. The first row shows the ability to recover weak
    boundaries. This sensitivity is not desirable in some cases as it leads the over-segmentation of the 2nd
    row. The third row suggests that the over-segmentation can be prevented by a stronger filtering but would also lead to
    a lower shape detection.}
\label{fig.watershed}
\end{figure}

\Cref{fig.watershed} shows the strength of the watershed to recover the boundaries of objects even on weak edge
responses that would be lost by thresholding the EPM. This is especially visible in the first row where the boundaries
of ``Place du Châtelet'' are leaking; nevertheless they are recovered in the segmentation. On the downside, this ability
to recover weak edges is also a bottleneck that can create false-boundaries as shown in the middle row where the place
around ``Eglise Notre-Dame'' is over-segmented because of some detection noise.

The filtering parameters (dynamic $h$ and area $\lambda$) are important to control the trade-off between the fact we
want to recover small/leaking regions (somewhat related to the \emph{recall}) and the false-detection of boundaries
(somewhat related to the \emph{precision}). This is illustrated with two sets of parameters A and B where B has more
restrictive filtering parameters. The third row of~\Cref{fig.watershed} shows that B has less over-segmentation but in
the two first rows, it misses some boundaries.

The decision to merge objects is actually very context-dependant, as it does not depend on the size of the component,
neither its volume, nor its shape. The watershed ``does its best'' to create the missing boundaries and, at the moment,
we have not managed to find better rules (e.g., with extinction values of some attributes) to filter out the basins of the
watershed.

\section{Evaluation}
\label{sec:eval}
To assess the performance of the proposed approach, we conducted a series of experiments on a fully manually annotated map sheet.
We report here details about this dataset we created and used,
the experimental protocol as well as the calibration procedures we followed,
the metrics we designed and used,
and discuss some results.

\textbf{Dataset} %
Among the multiple map sheets of the collection of Paris atlases, our work focuses on the particular sheet representing a central area of the city from year 1925~\cite{Whurer1925}.
We encourage the reader to refer to the \textbf{supplementary material} of this paper for a full-size view of this image.
Indeed, such map sheets are large by nature and were digitized with high resolution,
resulting in a 8500$\times$6500 image for the area of interest.
\\%
We carefully annotated the original image by creating line vector information for each edge of each object of interest in the map.
It should be noted that only a subset map strokes should be kept
as many objets are not relevant for our current study:
underground lines and railways, for instance, should be discarded.
The resulting vector information was rastered to produce:
i)~a reference edge map (a small dilation was applied so the resulting edges have a thickness of 3 pixels);
ii)~a reference label map identifying each shape to be detected.
\\%
We divided the image into three disjoint subsets:
a training set (rows 0 to 3999);
a validation set (rows 4000 to 4999);
and a test set (rows 5000 to 6500).
These areas were divided into 228 disjoint tiles of 500$\times$500 pixels.

\textbf{Protocol} %
Our goal in the evaluation protocol we designed was the assess the impact of the watershed stage in our pipeline.
We compared compared the performance of a baseline system, without watershed,
with our proposed approach: the same baseline augmented by a watershed stage
(see \cref{fig.pipeline}).
\\%
The baseline (without watershed) consists in a deep edge detection stage using the BDCN network presented in \cref{sec:deep_edge_det}.
This stage produces an edge probability map (EPM) as previously explained.
The network was trained on the training set using the validation set as control set during training.
To generate closed shapes, we simply thresholded the EPM and extracted the connected components.
We selected the best performing threshold value (9) on the validation set for fair comparison.
\\%
The proposed approach (baseline plus watershed) consists in adding a joint 
filtering on area and dynamic of the EPM followed by a watershed.
This approach produces a label map, i.e. a usable set of closed shapes,
as detailed in \cref{sec:mm_extraction}.
We selected the best performing values for area ($\lambda$) and dynamic $h$ parameters on the validation set.
\\%
To avoid losing topological information during component labeling (baseline) or during watershed,
theses steps were performed on the full image (with training, validation and test sets merged)
but the performance indicators were computed exclusively on the test set by masking other areas.

\begin{figure}[tb]
  \begin{minipage}{0.3\textwidth}
    \centering
    \includegraphics[width=\textwidth]{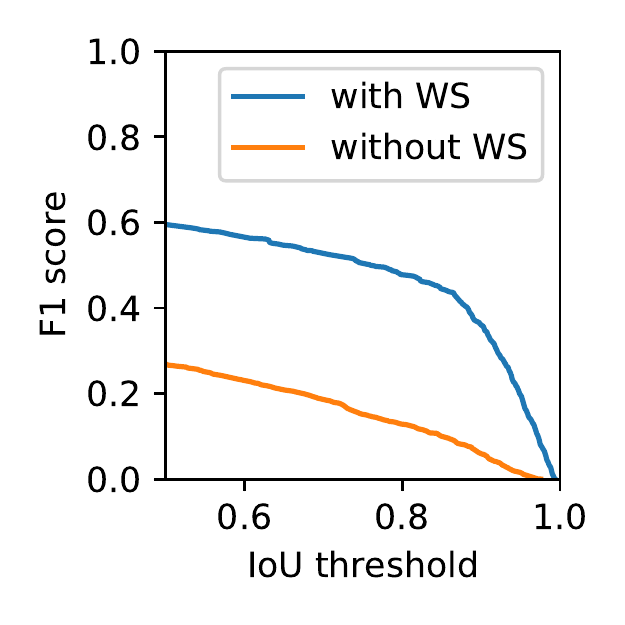}\hfill
  \end{minipage}%
  \hfill
  \begin{minipage}{0.7\textwidth}
    \centering
    \begin{tabular}{@{}l@{\quad}rrr@{\quad}rrr@{}}
      \toprule
      & \multicolumn{3}{c}{CC-labeling}
      & \multicolumn{3}{c}{Watersheding}
      \\
      \cmidrule(r){2-4}\cmidrule(){5-7}
      IoU & Precision &  Recall & F-score & Precision &  Recall & F-score \\
      0.50    &   0.20   &  0.39  &   0.27  &   0.74  &   0.50  &   0.59  \\
      0.80    &   0.10   &  0.19  &   0.13  &   0.60  &   0.40  &   0.48  \\
      0.90    &   0.04   &  0.09  &   0.06  &   0.45  &   0.30  &   0.36  \\
      0.95    &   0.01   &  0.02  &   0.02  &   0.25  &   0.16  &   0.20  \\
      \bottomrule
    \end{tabular}
  \end{minipage}%
  \caption{%
    \emph{Left}: comparison of the evolution of the shape detection F1-score across all possible IoU threshold with and without the watershed stage.
    \emph{Right}: evaluation metrics with and without watershed.}
  \label{fig:resultscurvetab}
\end{figure}

\textbf{Metrics} %
While it is common in segmentation challenges to evaluate the quality of object detection by
evaluating the precision and recall of edge detection at pixel,
such an approach would only evaluate the process halfway to our target application: closed shapes detection.
To evaluate shape detection, we need to identify pairs of matching shapes between a reference set ($R$) and a predictions set ($P$).
Because, in our particular case, shapes are disjoint among $R$ and also among $P$ (by construction),
we can leverage the following property:
as soon as the intersection over union (\emph{IoU}) between $r_i \in R$ and $p_j \in P$ is strictly superior to $0.5$,
then we know that no other element $r_k \in R, i \neq k$ can have a higher IoU with $p_j \in R$ than $r_i \in R$,
and reciprocally.
\\%
As soon as a pair of shapes $(r_i,p_j) \in R \times P$ exhibits $\mathrm{IoU}(r_i,p_j) = \mathrm{area}(\frac{r_i \cap p_j}{r_i \cup p_j}) = T > 0.5$
then we count a successful match under the threshold constraint $T$.
We introduce this threshold value so as to consider all possible values between 0.5 (exclusive) and 1 (inclusive)
and create a global indicator of the system under all potential quality requirements.
This allows us to count the number of 
correctly detected shapes (\emph{true positives} or $\mathrm{TP}$),
missed shapes ((\emph{false negatives} or $\mathrm{FN}$)),
and wrongly predicted shapes ((\emph{false positives} or $\mathrm{FP}$))
for every operating characteristics.
We derive from this set of measures two analysis tools.
\\%
First a precision ($\frac{\mathrm{TP}}{\mathrm{TP}+\mathrm{FN}}$), a recall ($\frac{\mathrm{TP}}{\mathrm{TP}+\mathrm{FP}}$) and a F1 score ($\frac{2\mathrm{TP}}{2\mathrm{TP}+\mathrm{FN}+\mathrm{FP}}$) curves
for all possible threshold values.
They offer a condensed view of the behavior of a system under all possible operating characteristics.
We suggest to use the area under the F1 score curve to compute a single performance indicator.
The values of this indicator ranges from 0 (worst) to 0.5 (best).
\\%
The second tool is a pair of visualization maps:
a precision map which associates for each predicted shape $p_j \in P$ the maximal IoU value $b_{pj}$ such as
$b_{pj} = \mathrm{argmax}_{r_i \in R}(\mathrm{IoU}(r_i, p_j))$,
and a recall map which associates for each expected shape $r_i \in R$ the maximal IoU value $b_{ri}$ such as
$b_{ri} = \mathrm{argmax}_{p_j \in P}(\mathrm{IoU}(r_i, p_j))$.
Each pixel of each shape is then assigned a color indicating the value of the maximal IoU:
red to yellow for values between 0 and 0.5,
and yellow to green for values between 0.5 and 1.
The darker the green the better the match in both maps.
The darker the red, the more serious the false positive (resp. negative) in precision (resp. recall) map.

\begin{figure}[tb]
  \centering
  \begin{tabular}{ccc}
    \includegraphics[trim={0 4cm 0 4cm},clip,width=0.485\linewidth]{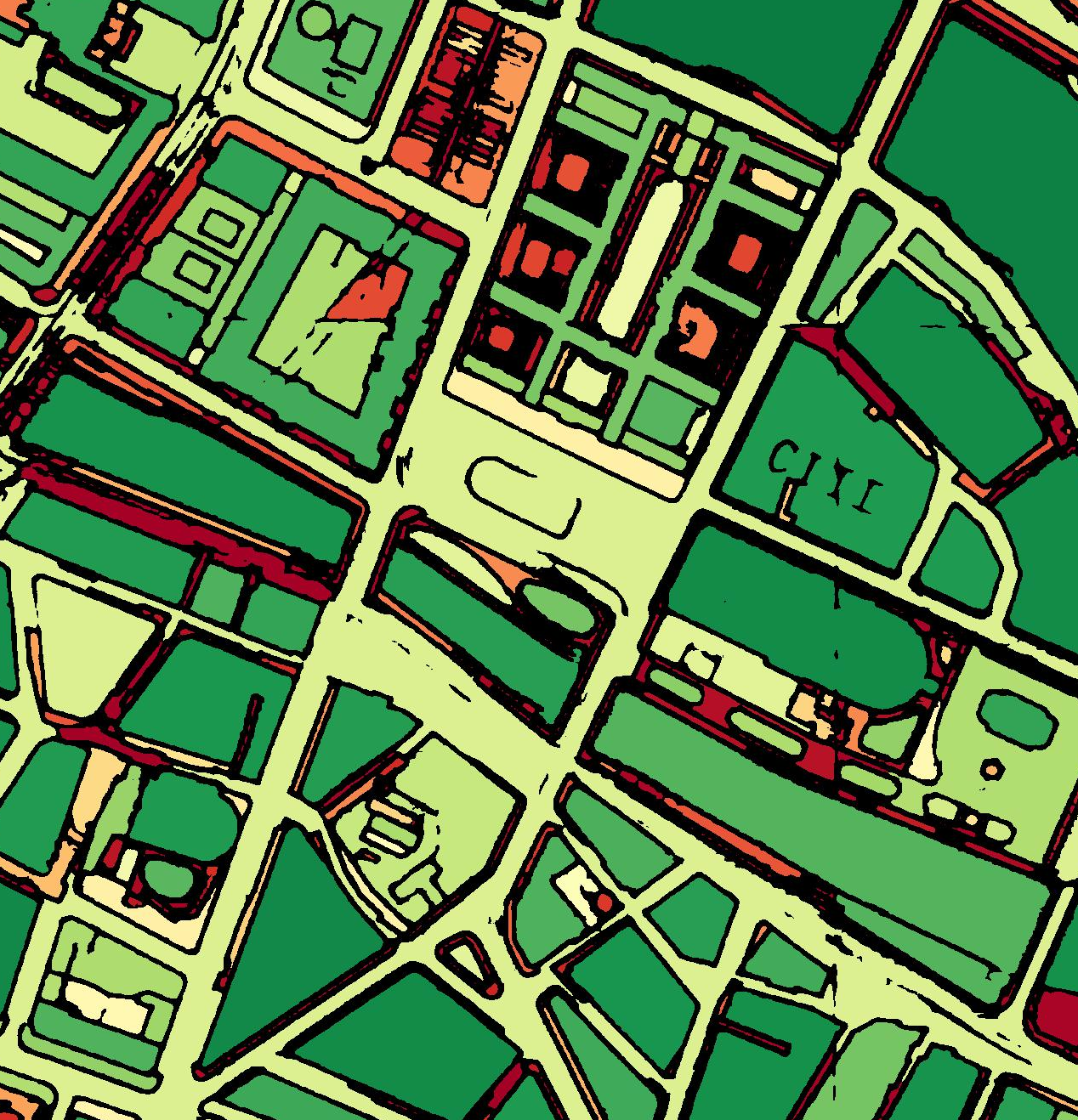} & 
    \qquad & 
    \includegraphics[trim={0 4cm 0 4cm},clip,width=0.485\linewidth]{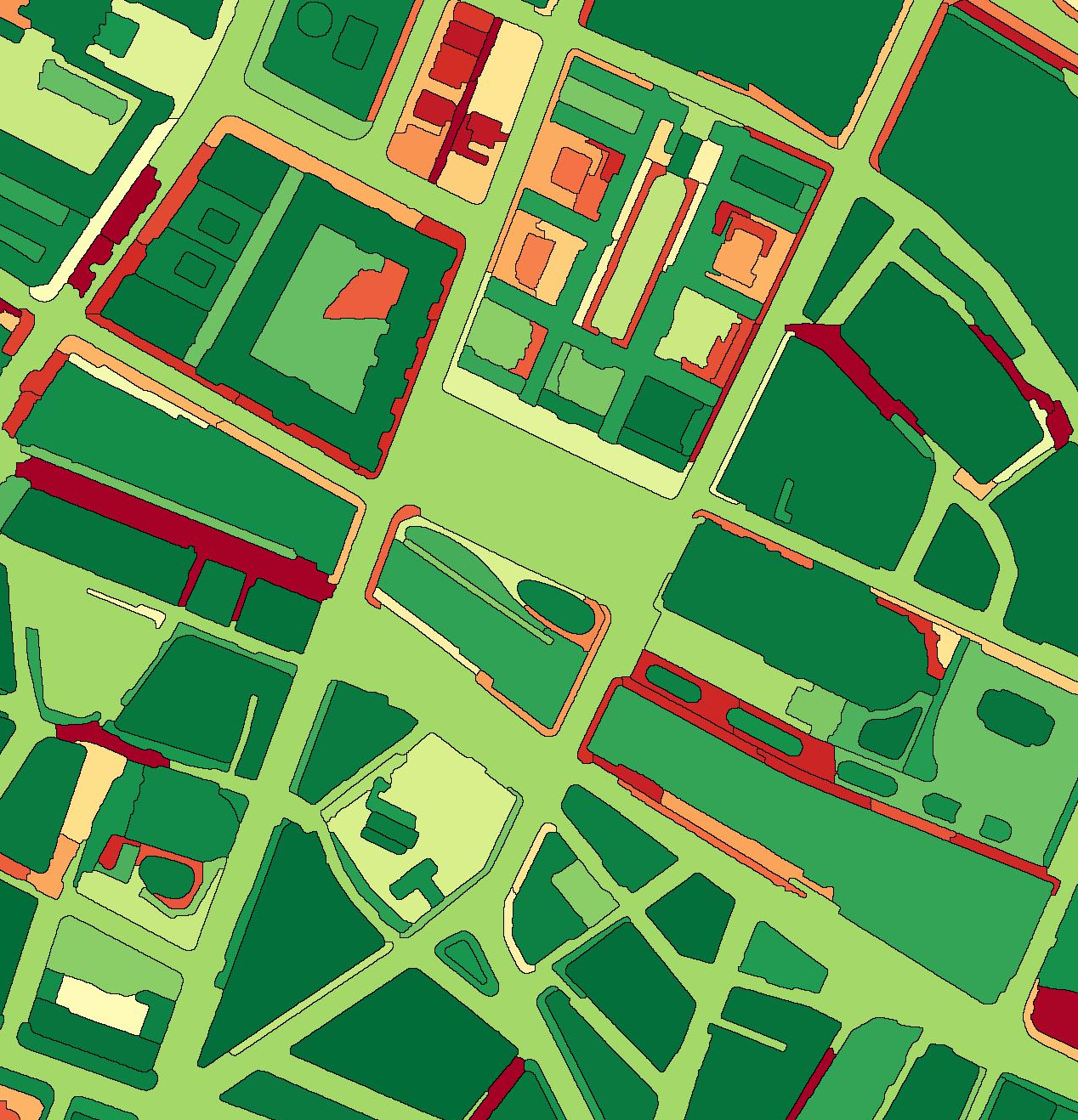}\\
    Precision map before watershed &
    &
    Precision map after watershed\\[0.5em]
    \includegraphics[trim={0 4cm 0 4cm},clip,width=0.485\textwidth]{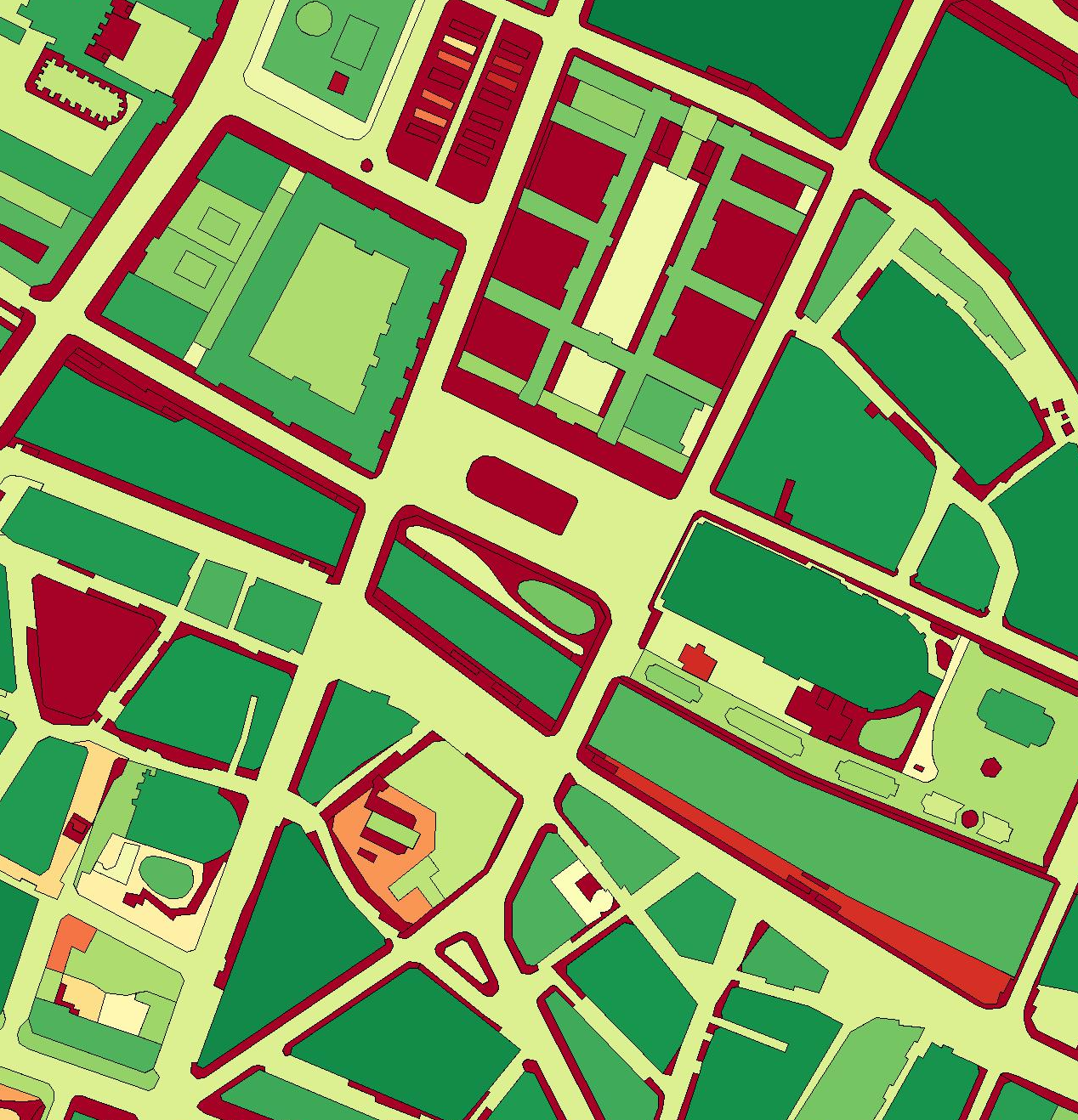}&
    &
    \includegraphics[trim={0 4cm 0 4cm},clip,width=0.485\textwidth]{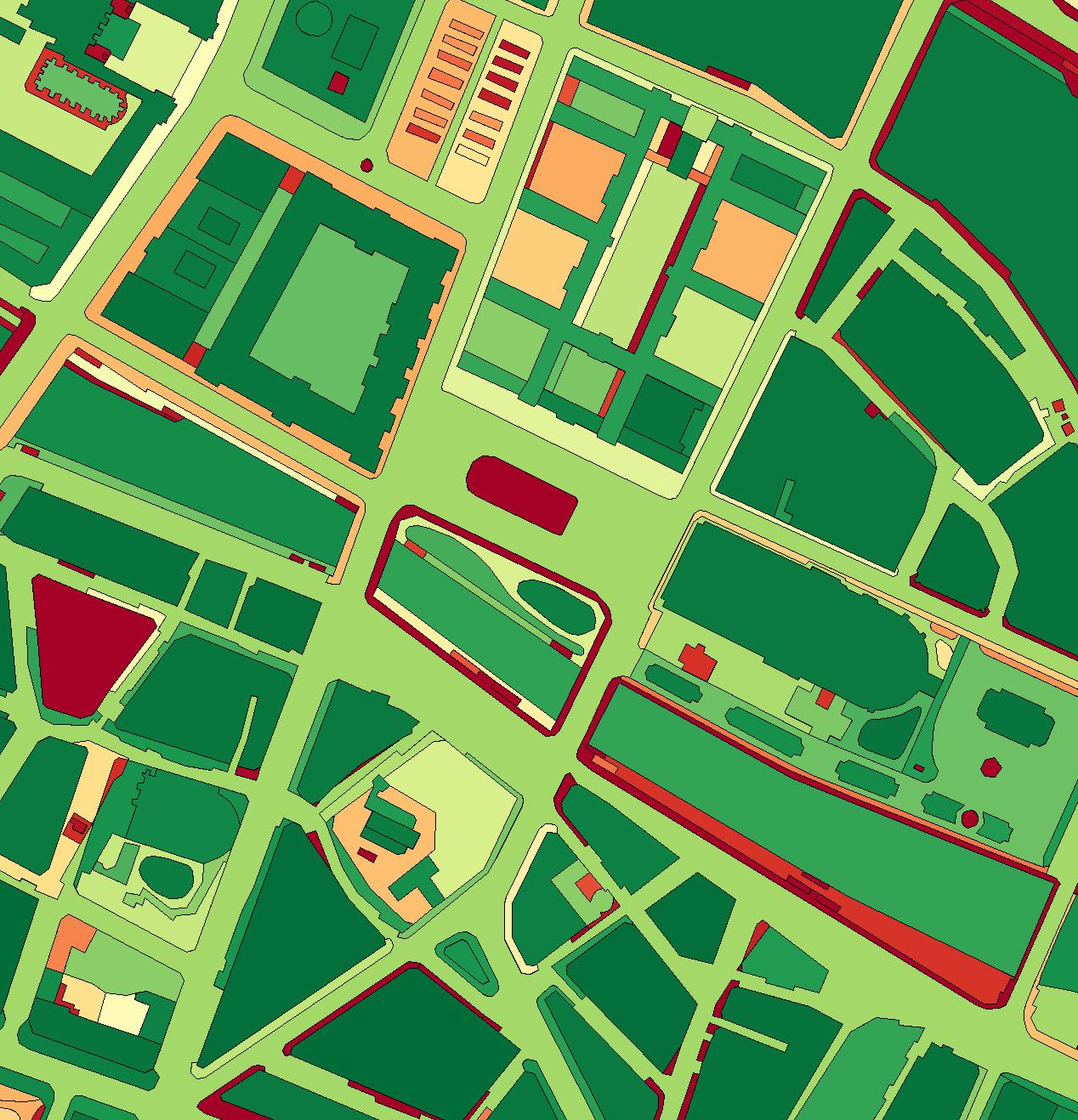}\\
    Recall map before watershed&
    &
    Recall map after watershed
  \end{tabular}
  \caption{Precision  and recall maps without and with watershed.}
  \label{fig:precisionrecallmaps}
\end{figure}

\textbf{Results and Discussion} %
We report here the results for the best calibrated variant 
of each of the two systems (baseline+connected component labeling vs baseline+watershed) under test.
\Cref{fig:resultscurvetab} (left) compares the evolution of the F1 score indicator for both systems under each possible IoU threshold.
\Cref{fig:resultscurvetab} (right) details the different indicators for several key values of IoU thresholds.
We can see from those results that the watershed post-processing consistently and significantly improves the quality of the results.
The precision and recall maps presented in \cref{fig:precisionrecallmaps} illustrate the benefits that the watershed post-processing bring to the deep edge segmentation:
it adjusts the border of the shapes (improves precision and recall);
it also removes small noise (improves precision);
and it also efficiently recovers some weak boundaries (improves recall).

\section{Conclusion}
In this paper, we propose an efficient combination of 
convolutional neural networks and mathematical morphology
to address the problem of closed shapes extraction in historical maps.
Convolutional neural networks (BDCN) allow us to efficiently detect edges while filtering unwanted features (text for instance).
Mathematical morphology is applied to the edge probability map created by BDCN to create closed shapes reliably.
The efficiency of our approach is shown by testing it on an open dataset.
We believe such a method will make the digitization process of historical maps faster and more reliable.

\bigskip
\noindent\textbf{Acknowledgements}\\
{
  \footnotesize
  This work was partially funded by the French National Research Agency (ANR):
  Project SoDuCo, grant ANR-18-CE38-0013.
}

\bibliographystyle{splncs04}
\bibliography{main-paper}

\end{document}